\documentclass[10pt,twocolumn,letterpaper]{article}

\usepackage{wacv}
\usepackage{times}
\usepackage{epsfig}
\usepackage{graphicx}
\usepackage{amsmath}
\usepackage{amssymb}
\usepackage{mydefs}

\def\wacvPaperID{****} %

\wacvfinalcopy %

\ifwacvfinal
\def\assignedStartPage{1} %
\fi

\ifwacvfinal
\usepackage[breaklinks=true,bookmarks=false]{hyperref}
\else
\usepackage[pagebackref=true,breaklinks=true,colorlinks,bookmarks=false]{hyperref}
\fi

\ifwacvfinal
\else
\fi

\begin{document}

\title{Video-Based Reconstruction of the Trajectories Performed by Skiers}

\author{Matteo Dunnhofer$^{\bullet}$\\
\and
Alberto Zurini$^{\bullet}$
\and
Maurizio Dunnhofer$^{\star}$
\and
Christian Micheloni$^{\bullet}$ \and
$^{\bullet}$Machine Learning and Perception Lab, University of Udine, Udine, Italy \\
$^{\star}$EYOF 2023 Organizing Committee, Amaro, Italy
}

\maketitle

\begin{abstract}
   Trajectories are fundamental in different skiing disciplines. Tools enabling the analysis of such curves can enhance the training activity and enrich the broadcasting contents.
However, the solutions currently available are based on geo-localized sensors and surface models. In this short paper, we propose a video-based approach to reconstruct the sequence of points traversed by an athlete during its performance. Our prototype is constituted by a pipeline of deep learning-based algorithms to reconstruct the athlete's motion and to visualize it according to the camera perspective. This is achieved for different skiing disciplines in the wild without any camera calibration.
We tested our solution on broadcast and smartphone-captured videos of alpine skiing and ski jumping professional competitions. The qualitative results achieved show the potential of our solution.
\end{abstract}

\section{Introduction}
An optimal trajectory is one of the key factors to achieve higher performance in different skiing disciplines. For example, in alpine skiing the turns performed by an athlete form a trajectory of points on the slope surface that if optimized with respect to the position of turn gates can lead to saved time during the descent \cite{trajski,cai2020trajectory}. In ski-jumping, a trajectory is formed by the points traversed during the flight phase and different flight paths influence the distance of the jumps \cite{hubbard1989multisegment,muller2006physics}. 
The optimization of these curves can lead to an increased chance of victory.
It is hence important to analyze the trajectory executed by the athlete in order to determine specific points of the performance that can be correlated to the final scores (e.g. overall time taken or distance jumped). Potential applications of such kind of trajectory analysis tool could be intelligent review systems that would enhance the training activities, but also richer broadcasting contents that would increase the engagement of spectators.

To reconstruct the trajectory of athletes in winter sport disciplines, the current standard practice \cite{gilgien2013determination,kruger2010application} is to put sensor devices (e.g. GNSS trackers, IMUs) on the body or skis and use a precise surface model to map the position and data coming from such sensors on the surface model at the different time steps. 
The drawback of these approaches is that they require the careful  and time-consuming installation of the sensor devices on the athlete and the acquisition of a precise ground model. Furthermore, such an approach could be not always achievable during competitions.
Computer vision techniques applied on videos capturing the athlete's performance is a valid option to obtain trajectories without the need of sensor networks nor ground surface models. The benefit of a video-based approach is even more evident considering the usual practice of video reviewing during training or the amount of video material  produced by the broadcasting of competitions.
Vision-based techniques have been successfully used in other sport disciplines to reconstruct the trajectory of various kinds of ball \cite{Chen2011,kotera2019intra} and player movements \cite{Calandre2021,chen2018player}.
However, to the best of our knowledge, no study is currently present to reconstruct the trajectory of skiing athletes in videos.

In this short paper, we present a prototype to
achieve the reconstruction of the trajectory of skiers in videos acquired from uncalibrated and unconstrained cameras.
Our algorithm works online, i.e. takes in input the latest frame of a streaming video and outputs the trajectory executed by the athlete in the previous time steps with the correct perspective with respect to the scene appearing in that frame.
The solution first runs a visual tracker \cite{Stark} to follow the target skier across all the previous frames up to the latest. Then, a key-point detection and matching algorithm \cite{SuperPoint,SuperGlue} is employed to estimate the motion of static key-points across the consecutive frames. The matched key-points are given to a RANSAC-based algorithm \cite{RANSAC,degensac} to estimate the homography representing the perspective transformation between those. Such a transformation is used to map all the points traversed by the athlete to the correct perspective, achieving the reconstruction of the trajectory with respect to the camera movements and ultimately giving a 3D effect.
The performed qualitative tests on broadcast and handheld camera videos of alpine skiing and ski-jumping show the potential of the proposed solution. Further research is needed to make this idea applicable in practice. We point out possible future research directions.

\section{Methodology}
\label{sec:method}

\subsection{Preliminaries}
The videos given as input to our solution are considered to capture the performance of an individual athlete while he/she is constantly visible in the scene. 
We do not put any constraints on the configuration (intrinsic and extrinsic parameters) of the camera that captured the videos.
More formally, we consider a video
$\video = \big\{ \frame_t \in \images \big\}_{t=0}^{T}$
as a sequence of frames $\frame_t$, where $\images =  \{0,\cdots,255\}^{w \times h \times 3}$ is the space of RGB images and $T \in \mathbb{N}$ denotes the number of frames.
We use $p_t = (x_t, y_t)$ to denote the coordinates of the point that summarizes the position of the athlete in the image coordinate system (e.g. the point of contact between the athlete and the ground surface).
The goal of our system is to produce a trajectory $\traj_t = \{ \point_i \}_{i=0}^{t-1}$ which is the sequence of points traversed by the athlete %
in the 3D environment mapped in the 2D space of $\frame_t$.

\begin{figure}[t]
\centering
  \includegraphics[width=.65\columnwidth]{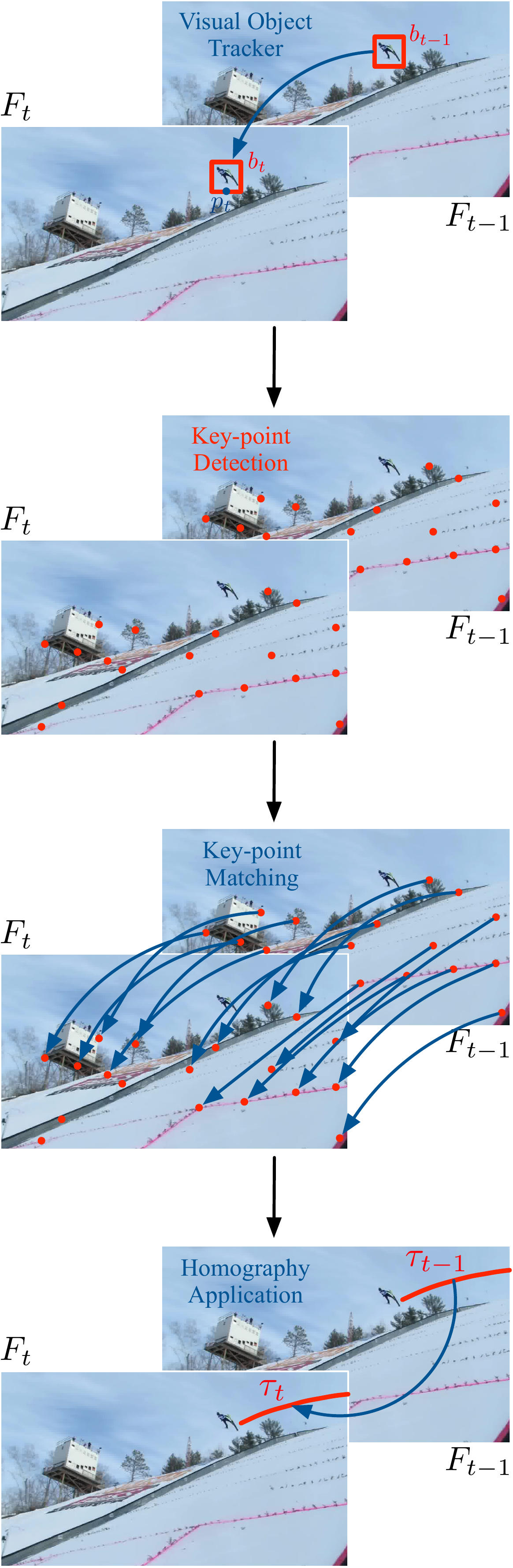}
  \caption{Schematic visualization of the main steps performed by our solution to obtain the trajectory $\tau_t$ at each time step $t$ using the consecutive frames $\frame_{t-1}, \frame_t$.} 
  \label{fig:pipeline}
\end{figure}

\subsection{Pipeline}
Figure \ref{fig:pipeline} presents a schematic representation of the pipeline constituting the proposed trajectory reconstruction algorithm.
The solution works in an online fashion. This means that at every $\frame_t$ the only available information to produce the trajectory $\tau_t$ is contained in $\frame_t$ and in all the preceding frames. 
This setting makes the solution suitable for real-time applications since it does not require waiting for the athlete's execution to be terminated for the trajectory to be produced. Moreover, our algorithm is general and it can be applied to different disciplines without specific tuning.
The proposed method processes each $\frame_t$ sequentially. $\frame_t$ is first given to a visual object tracking algorithm designed to model the motion of the athlete and to provide its position $\point_t$ in the latest frame. Then, the solution estimates the homography transformation $\homo \in \mathbb{R}^{3\times3}$ existing between $\frame_t$ and $\frame_{t-1}$. This is achieved by finding and matching particular image key-points present in $\frame_t$ and $\frame_{t-1}$ and using a RANSAC-like algorithm on top of the matchings to find $\homo$. 
Considering that in individual winter sport the environment of the course is generally composed of static objects (e.g. banners, line markers, etc.), it can be advantageous to compute their displacement in consecutive frames to quantify the camera motion.
$\homo$ is used to map the points of the trajectory $\tau_{t-1}$ for the previous frame into the coordinate system of the current frame, thus obtaining the trajectory $\tau_{t}$. After that, $\point_t$ is appended to $\tau_t$ to obtain all the points traversed by the athlete with respect to $\frame_t$.
We now describe the different components of the algorithm in more detail.

\begin{figure*}[t]
\centering
  \includegraphics[width=\linewidth]{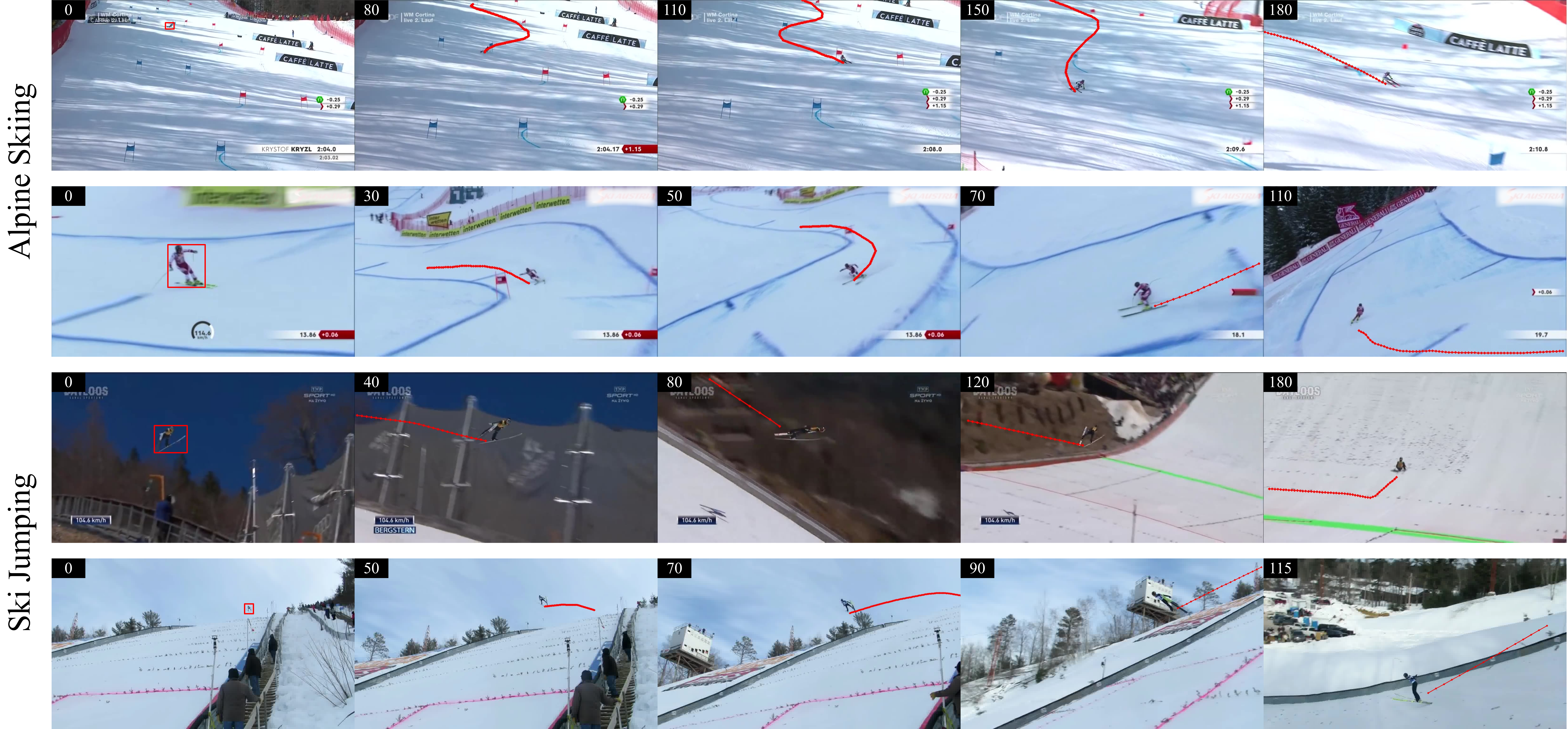}
  \caption{Examples of the trajectory produced by our solution. The number in the top-left corner of each image reports the frame index $t$ in the video. In the first frame, the bounding-box $\bbox_0$ is reported by the red rectangle. In the other frames, the red track is the trajectory $\tau_t$ reconstructed by our pipeline for that frame. The first two rows of frames show examples of the algorithm applied on broadcasting videos of giant slalom and downhill skiing. The last two rows show applications to a broadcasting video and a smartphone video of ski jumping.} 
  \label{fig:trajs}
\end{figure*}

\paragraph{Athlete Tracking.}
The first step of the pipeline is to exploit a visual object tracker \cite{Dunnhofer2019,Dunnhofer2020accv} to track the motion of the athlete across all the frames up to $\frame_t$. We used a tracker outputting a bounding-box $\bbox_t = (x^{(\bbox)}_t,y^{(\bbox)}_t,w^{(\bbox)}_t,h^{(\bbox)}_t) \in \mathbb{R}^4$ at every $\frame_t$. The $x^{(\bbox)}_t,y^{(\bbox)}_t$ represent the coordinates of the top-left corner of the box while $w^{(\bbox)}_t,h^{(\bbox)}_t$ are employed to get an estimate of the width and height of the athlete's appearance. We consider the position of the athlete $\point_t = (x_t, y_t)$ as $x_t = x^{(\bbox)}_t + \frac{w^{(\bbox)}_t}{2}$ and $y_t = y^{(\bbox)}_t + h^{(\bbox)}_t$. Given an accurate bounding-box, $\point_t$ represents the closest point to the contact between the athlete's feet and the ground. 
For disciplines in which the athlete lies constantly on the ground, such a setting allows to estimate the point traversed by the athlete.
The tracker is initialized in the first frame $\frame_0$ of the video with the bounding-box $\bbox_0$ that outlines the appearance of the target. Such a piece of information can be obtained by asking a human operator to provide the bounding-box for the athlete of interest via some user-friendly annotation system or by a specific athlete detection algorithm. %
We used the state-of-the-art deep learning-based method STARK \cite{Stark,VOT2021} (with pre-trained parameters) as visual tracker because of its ability in providing bounding-boxes that fit accurately the appearance of a large variety of target objects.

\paragraph{Frame Matching.}
The tracker allows to model the motion of the athlete in each $\frame_t$. 
We want
to render such motion in relation to the perspective of the scene and the athlete's execution, ultimately giving a 3D effect. To achieve this we use the homography matrix $\homo$.
The first step to compute the $\homo$ between $\frame_{t-1}$ and $\frame_t$ is to run an image key-point detector to obtain significant points of interest in the field of views of both frames. For this task, we employed the deep learning-based methodology SuperPoint \cite{SuperPoint} because of its state-of-the-art performance. Particularly, we used the pre-trained instance of the algorithm optimized for outdoor scenarios provided by the authors \cite{SuperPoint}. From the sets of key-points, we excluded those located within the bounding-box $\bbox_t$ because they belong to a non-static object. In the case of broadcasting videos, we also discarded all the key-points lying on the superimposed banners showing the characteristics of the athlete's performance (e.g. running time).
Once the key-points have been determined, the matching algorithm is executed to find those key-points that correspond to the same visual features in $\frame_t$ and $\frame_{t-1}$. This allows to obtain an alignment between the same points on the images that expresses how the static objects have moved between the frames. To perform the matching we exploited the SuperGlue algorithm \cite{SuperGlue} which is a graph-based deep-learning method that focuses on the global organization of key-points in order to find matches between them. As for SuperPoint, we used the pre-trained instance of the algorithm optimized for outdoor scenarios as provided by the authors \cite{SuperGlue}.

\paragraph{Homography Estimation.}
With the alignments of matched key-points, we are able to obtain the homography matrix $\homo$.  
This is achieved through an instance of the DEGENSAC algorithm \cite{degensac} which applies an iterative optimization procedure on the matchings in order to find the best homography matrix that explains them.
We found DEGENSAC to work better than a standard RANSAC instance. 

\paragraph{Homography Application and Trajectory Reconstruction.}
After that the homography is determined, it is used to map the points of the previous trajectory $\traj_{t-1}$ in the new frame. 
In more detail, at each frame $\frame_t$, $\traj_{t-1}$ consists of all the points given by the visual object tracker in the preceding $t-1$ frames and localized according to the perspective of the previous frame $\frame_{t-1}$. The trajectory $\traj_t$ for $\frame_t$ is obtained by the multiplication of each $\point_i \in \traj_{t-1}$ by the homography matrix, i.e. $\traj_t = \{ \point_i \}_{i=0}^{t-1}, \point_i = \point_i \cdot \homo$. Then, $\traj_t$ is also appended $\point_t$ given by the tracker for $\frame_t$ which represent the latest position of the athlete. 
At the first frame in which the frame matching step is executed $\frame_1$, $\traj_{0} = \{ \point_0 \}$ is composed only of the point extracted by the bounding-box which highlights the target athlete.
After its reconstruction, spline interpolation is also applied to $\traj_t$ to make the trajectory smoother.

\begin{figure}[t]
\centering
  \includegraphics[width=\columnwidth]{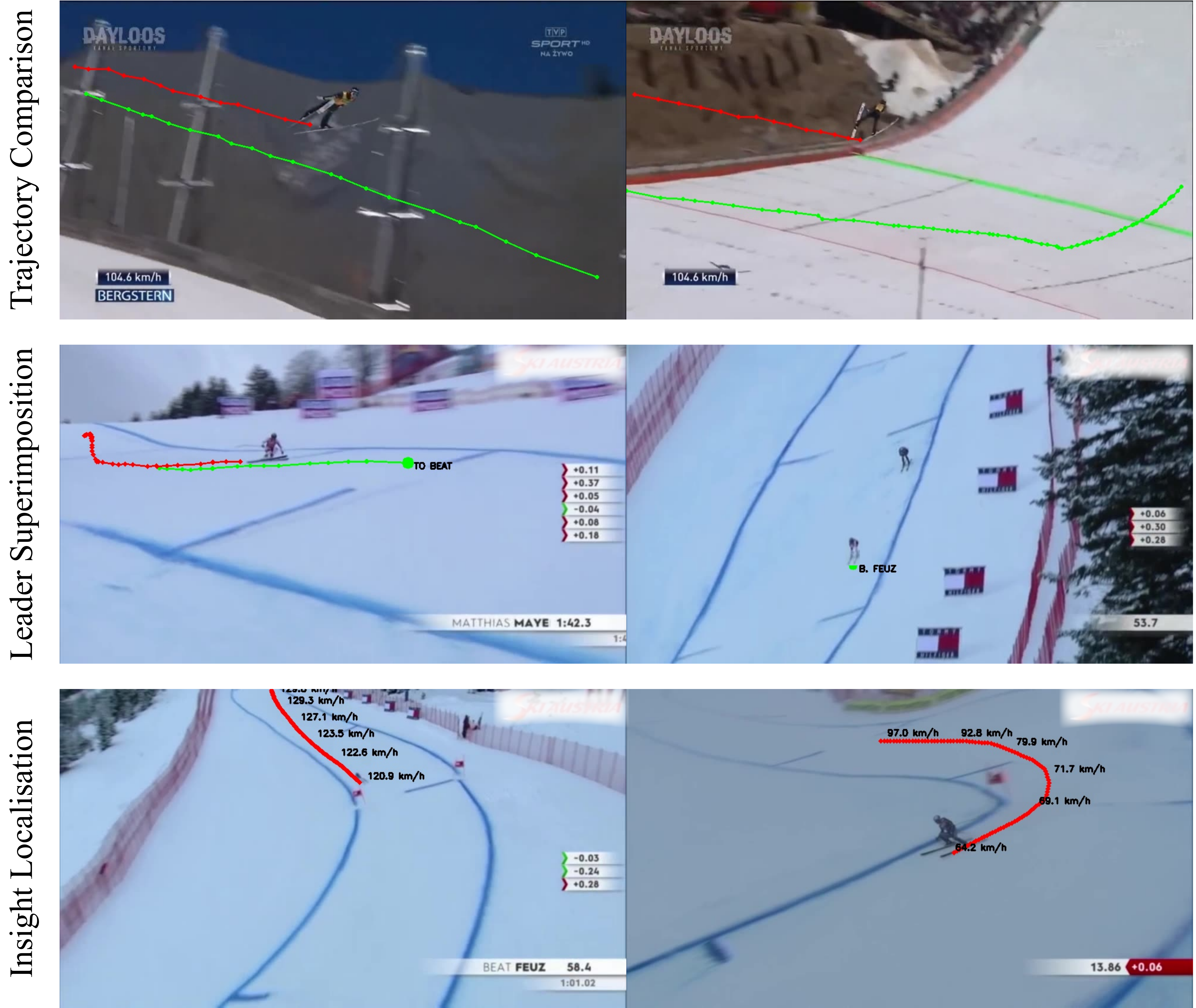}
  \caption{Examples of analytical applications feasible with our solution. The first row of images reports two different frames in which the flight trajectory of an athlete (red track) is compared with the flight trajectory of another jumper (green track). The second row reports two different scenarios in which the trajectories of two skiers are synchronized in time in order to visually report their spatial gap. The third row presents how  the computed trajectory can be augmented to give localized insights about the performance of the athlete (e.g. by using information about the speed coming from IMU sensors worn by the skiers). } 
  \label{fig:apps}
\end{figure}

\section{Experiments and Discussion}
We performed qualitative experiments on our prototype. This is due to the non-availability of public datasets suitable for the evaluation of trajectory reconstruction in winter sports applications. Future work will be dedicated to build an accurate set of videos for quantitative validation.
We tested our solution for the reconstruction of the trajectory executed by alpine skiers while skiing by ski jumpers while flying. 
We acquired videos of such two disciplines on YouTube. In particular, for alpine skiing, we tested our solution on broadcast videos of the giant slalom taken place at the FIS Alpine Ski World Championship in Cortina 2021 and of the FIS Alpine World Cup downhill race in Kitzb\"uhel 2021. For ski jumping, we ran our solution on broadcast videos of the FIS Ski Flying World Cup competition in Planica 2019 and on videos acquired by smartphones during the FIS Ski Jumping Continental Cup in Iron Mountain 2020.
No specific adaption for the two settings was performed.

Figure \ref{fig:trajs} shows examples of the performance achieved by our solution. The trajectories produced are consistent with the past motion of the athlete, and the reconstruction capability seems to be robust to the different camera movements, to the blurred background, and to the changes in illumination conditions. Overall we think our solution to be promising.
The pictures in Figure \ref{fig:apps} present some particular analytical applications based on our solution. The first row of images reports two frames in which the trajectory of the jumper (in red) is compared with the trajectory of another jumper (in green).
The second row shows two different visualizations of the insertion of the trajectory (left image, green trajectory) and the visual appearance (right image, highlighted by the green dot) of the competition's leader. These kinds of solution have been achieved by synchronizing the videos of the two athletes and computing a homography between the time-paired frames by the frame matching procedure described in Section \ref{sec:method}.
The third row displays two images in which the trajectory is augmented with insights about the athlete's performance. In this case, the speed data obtained by IMU sensors worn by the skiers and synchronized with the video frames.

Further work is needed to make this solution effective. First, the error committed in the reconstruction of the trajectories should be quantified using labeled data. For example, the displacement in centimeters with respect to the true trajectory performed by the athlete could be a valuable measure of the precision of the proposed solution.

We hypothesize that the performance of the system could be improved by better integrating the different modules of the pipeline, and potentially through an end-to-end optimization stage of the learning modules and backbone networks involved.
The system could be also enhanced by exploiting human pose trackers instead of bounding-box ones. Indeed, a human skeleton-based tracker should provide a better and more precise localization of the body of the target skier. Such a representation could be exploited to compute a more consistent point of contact between the athlete and the snow surface. Furthermore, the motion modeling of the different human body parts could enable the development of solutions able to simultaneously reconstruct the trajectory of disparate parts of the athlete (e.g. hands or feet). A similar idea could be also exploited to compute the pose trajectory of the single skis if a pose estimator/tracker for this kind of object \cite{SkiPose} is used. 

\begin{figure}[t]
\centering
  \includegraphics[width=\columnwidth]{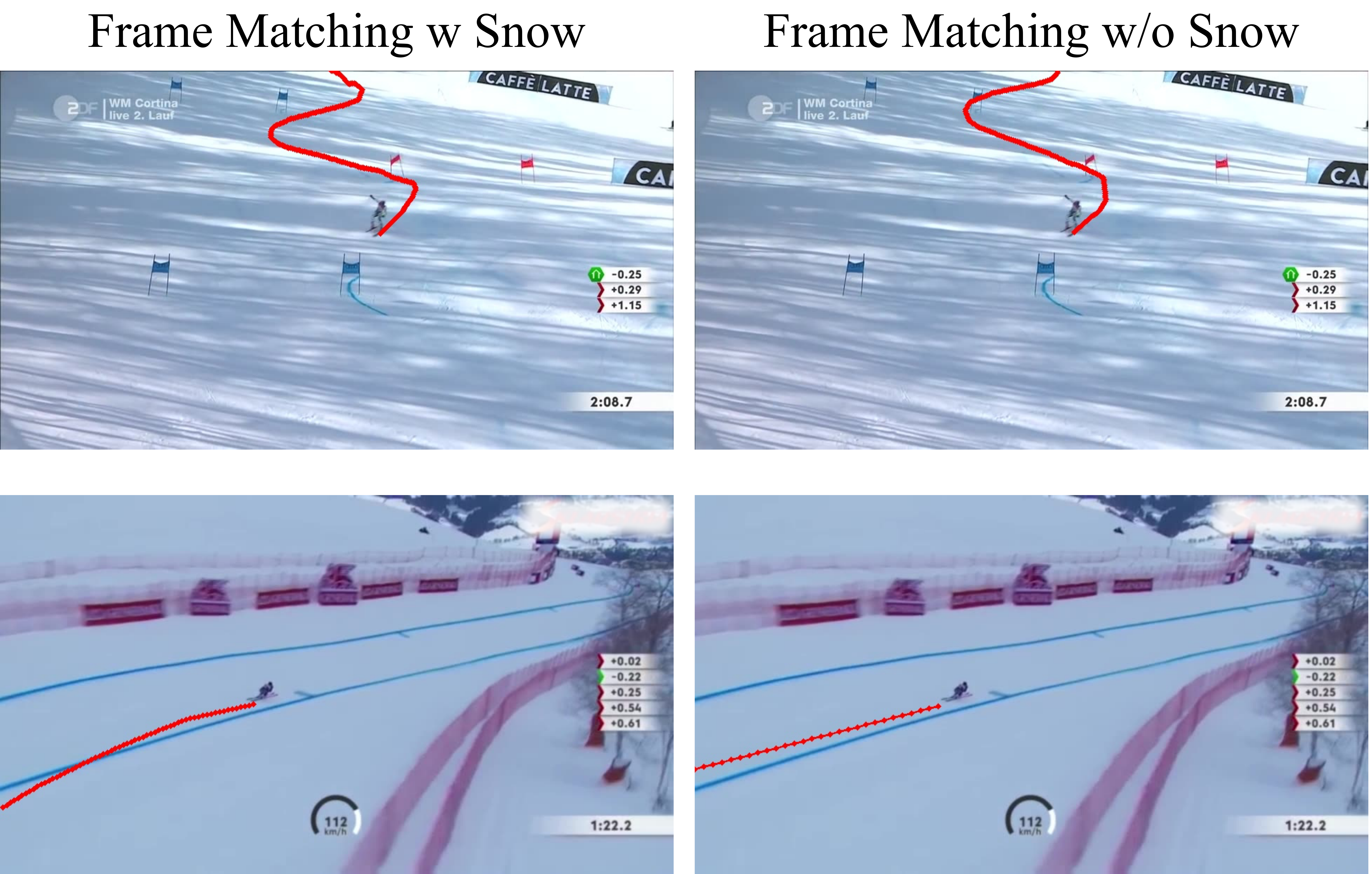}
  \caption{Examples of the influence of the snow texture on the reconstruction of the trajectory. The two rows of images show the same frame of two different videos. The frames of the first column report the trajectory obtained with the homographies computed using all the original key-points detected by SuperPoint (and then matched by SuperGlue). The second column shows the frames in which the SuperPoint-detected key-points lying in image locations where the snow is present are filtered out before matching. Such an operation enables the estimate of a more consistent homography, ultimately resulting in the better reconstruction of the trajectory.} 
  \label{fig:snow}
\end{figure}

Finally, we think that the better exploitation of the specific cues appearing on the slope and in training/competition scenarios could lead to an enhanced trajectory reconstruction performance. Indeed, as showed in Figure \ref{fig:snow}, in some of our experiments, we found that the snow texture provided no useful information for key-point detection. This issue influenced the homography estimation and ultimately led to wrong trajectory reconstructions. Filtering those key-points lying on image positions with whitish appearance -- hence matching only key-points belonging to other visual features (e.g. line markers, banners, etc.) -- allowed a better estimate of the homography and consequently an improved trajectory reconstruction.

{\fontsize{8.5}{9.5}\selectfont
\bibliographystyle{ieee_fullname}
\bibliography{egbib}

\end{document}